\documentclass[acmtog]{acmart}

\newcommand{\norm}[1]{\left\lVert#1\right\rVert}
\renewcommand{\sectionautorefname}{Section}
\let\subsectionautorefname\sectionautorefname 
\let\subsubsectionautorefname\sectionautorefname

\newcommand{\edit}[1]{\textcolor{blue}{#1}}
\renewcommand{\edit}[1]{#1}

\newenvironment{editenv}{
    \bgroup\color{blue}
    \captionsetup{labelfont={color=blue},textfont={color=blue}}
}{
\egroup
}

\newcommand{\bfpara}[1]{\paragraph{#1}}

\AtBeginDocument{%
  \providecommand\BibTeX{{%
    \normalfont B\kern-0.5em{\scshape i\kern-0.25em b}\kern-0.8em\TeX}}}

\setcopyright{acmlicensed}
\acmJournal{TOG}
\acmYear{2020}
\acmVolume{39}
\acmNumber{4}
\acmArticle{40}
\acmMonth{7}
\acmDOI{10.1145/3386569.3392422}

\acmSubmissionID{303}

\begin{document}

\title{Character Controllers Using Motion VAEs}

\author{Hung Yu Ling}
\affiliation{
  \institution{University of British Columbia}
  \city{Vancouver}
  \country{Canada}
}
\email{hyuling@cs.ubc.ca}

\author{Fabio Zinno}
\affiliation{
  \institution{Electronic Arts Vancouver}
  \city{Vancouver}
  \country{Canada}
}
\email{fzinno@ea.com}

\author{George Cheng}
\affiliation{
  \institution{Electronic Arts Vancouver}
  \city{Vancouver}
  \country{Canada}
}
\email{gecheng@ea.com}

\author{Michiel van de Panne}
\affiliation{%
  \institution{University of British Columbia}
  \city{Vancouver}
  \country{Canada}
}
\email{van@cs.ubc.ca}

\renewcommand{\shortauthors}{Ling, H. et al}

\begin{abstract}

A fundamental problem in computer animation is that of
realizing purposeful and realistic human movement given a sufficiently-rich set of motion capture clips.
We learn data-driven generative models of human movement using autoregressive conditional 
variational autoencoders, or Motion VAEs. The latent variables of the learned autoencoder define the action space 
for the movement and thereby govern its evolution over time.  
Planning or control algorithms can then use this action space to generate desired motions.
In particular, we use deep reinforcement learning to learn controllers
that achieve goal-directed movements.  We demonstrate the effectiveness of the approach on multiple tasks.
We further evaluate system-design choices and describe the current limitations of Motion VAEs.

\end{abstract}



\begin{teaserfigure}
  {\centering \includegraphics[width=0.8\textwidth]{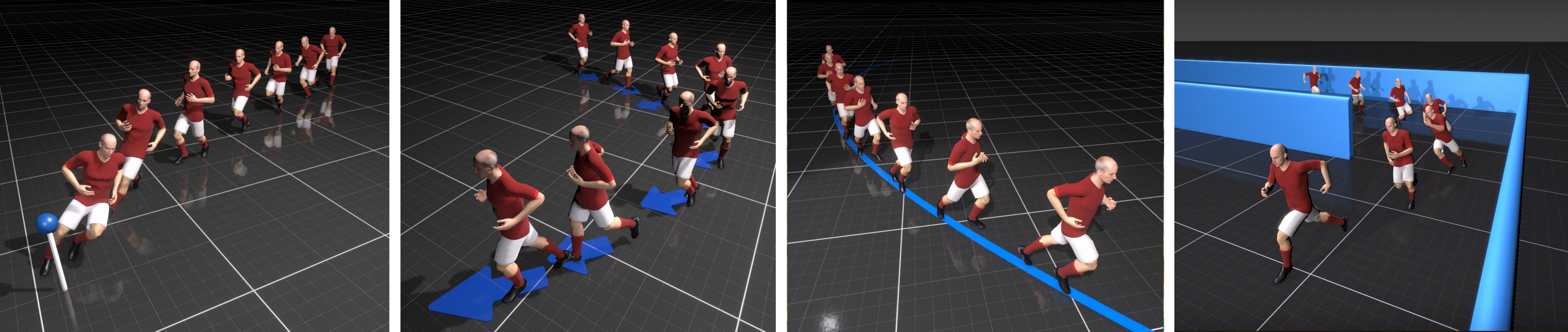}\\}
  \caption{
  Given example data, we learn an autoregressive conditional variational autoencoder (VAE) that predicts the next pose one frame at a time.  A variety of task-specific control policies can then be learned on top of this model.}
  \label{fig:teaser}
\end{teaserfigure}

\maketitle

\section{Introduction}

Given example motions, how can we generalize these to produce new purposeful motions?
This problem is at the core of interactive character animation and control,
with applications that include games, simulations, and virtual reality.
The solutions should ideally produce high-quality motion, be compact, 
be efficient to compute (at runtime), and support a variety of goal-directed behaviors.

In this paper, we take a two-step approach to this problem. 
First, we learn a kinematic generative model of human motion given the example motion data.
This is based on an {\em autoregressive conditional variational autoencoder}, which we 
refer to more simply as a {\em motion VAE (MVAE)}. Given the current character pose,
the MVAE predicts the pose at the next time step.  Importantly, the model can 
produce a {\em distribution} of next-state predictions because it is also conditioned
on a set of stochastic latent variables -- each sampling of these variables
corresponds to a different feasible next-state prediction.
The model is autoregressive, meaning that the current predicted pose 
becomes the current character pose for the following prediction.

Given a trained MVAE, it can be controlled to generate desired motions in several ways.
The simplest is to randomly sample from the next-state predictions at each time-step,
which produces a random walk through the learned dynamics of the MVAE.
More interestingly, we can treat the stochastic variables that govern the
next-state predictions as the action space for a reinforcement learning problem, \edit{which is not possible for learning approaches based purely on supervised learning}.
Given a reward function that defines the goals of the character, a control policy
can then be learned which uses this action space to guide the 
generative model in accordance with those goals.

We note that VAEs have been previously identified as a promising class of models for kinematic motion generation,
along with RNNs.  However, the stable generation of long motion sequences is
commonly acknowledged as a significant challenge, as are issues of motion quality.
MVAEs produce high-quality results as demonstrated with fully-skinned characters and with no additional foot-skate cleanup. 
Further, unlike previous approaches using
a memory-enabled RNN architecture, we show that the MVAE can work well in a memoryless fashion,
i.e.,  conditioned only on the previous pose (which includes velocities).
The MVAE decoder architecture plays an important role with regards to motion quality and stability.
We provide insights with respect to generalization and over-fitting, as well
as documenting other remaining limitations of the MVAE architecture.
Importantly, we show the usability of MVAEs in an RL setting and 
demonstrate how an energy penalty can be used to better model human motion preferences.

Our contributions are as follows:
\begin{itemize} 
\item We introduce a generative VAE motion model, the MVAE, capable of producing stable high-quality human motions.
   We document the key algorithmic and architectural features that are needed to do this successfully.
\item We show that reinforcement learning can effectively use the MVAE generative
   model to produce compact control policies. These can then be coupled with the MVAE model to produce desired goal-directed movement.
\end{itemize}

\section{Related Work}
\label{sec:related}
\newcommand{\direct}{direct prediction}
\newcommand{\Direct}{Direct prediction}
\newcommand{\indirect}{model-then-control}
\newcommand{\Indirect}{Model-then-control}

The animation problem being solved is that of data-driven time-series prediction,
additionally conditioned on the desired goals of a movement.
We can further characterize solutions according to additional attributes:
(i) {\em kinematic} vs.\ {\em physics-based}: the former directly predicts future motion without regard 
for physics, while the latter uses a physics simulation to generate movement.
(ii) {\em \direct} vs.\ {\em \indirect}: if the example data
is already considered to be moving in accordance with the desired task goals, the predictive policy can be
trained directly using supervised learning. Otherwise, we describe it as being an indirect approach,
where the example data is first used
to learn a dynamics model, and a method such as reinforcement learning is then used to learn
a control policy on top of this model to realize the motions for the desired task.
(iii) {\em non-parametric model} vs.\ {\em parametric model}:  parametric models, such as neural networks,
discard the original motion data, while non-parametric models keep the original data.
In what follows, we focus heavily on kinematic motion synthesis methods,
and we further structure our review of related work according to the remaining attributes.
Locomotion is most frequently used as the desired task to solve.
For a broader survey on kinematic character animation methods, we refer the reader to~\cite{CGF-2010-survey}.

\subsection{Kinematic motion synthesis}

\bfpara{\Direct, non-parametric models}
In this category, the original example data is considered to directly embody example solutions to
the desired task, and the data is directly used to construct the generated motion.
The simplest version uses manually-designed control logic
that plays specific clips in specific situations, 
with possible multi-way blends for interpolation between two clips being played in parallel,
and to enable seamless-transitions between clips.
Instead of working with motion capture clips, {\em motion matching} \cite{MotionMatching} works with 
example motions at the level of individual frames. At run-time, it seeks to choose the best
possible next frame from a database of motion capture data when given the previous pose.  
The locomotion task itself is embedded in the feature vector used in a $k$-NN query.
This vector contains root information from the past, so as to be able to 
find a match that is compatible with the ongoing movement, and also root information
from the future, so as to be compatible with the desired motion as demanded by the task. 
Limitations include the generated behaviors being sensitive to 
the choice of time window and feature weighting that are used for the matching process.
Embedding the task in the feature vector also makes it difficult to apply this framework to non-locomotion tasks.

\bfpara{\Indirect, non-parametric models}
A different approach is to first use the example motions to learn or develop a model of the space of possible motions
and how they connect. This model can then be used by a planning or reinforcement learning algorithm
which, for a given character state in the world, determines an optimal traversal path through the model
to achieve a desired behavior or goal, as defined by the task.
Motion graphs~\cite{MotionGraphs} were an early method of explicitly inferring the connectivity
between arbitrary frames on motion clips, based on a pose distance metric being less than
a given threshold. It can also be useful to segment the motion into short clips with the use of
constraint frames, in a way that allows the construction of a valid animation from any sequence of these clips
without foot skating.  This creates an implicit fully-connected motion graph. 
While motion graphs have an explicit enumeration of the transition possibilities,
some transitions will still be smoother and more natural than others.
This can still be taken into account via an additional transition reward while planning a movement on the motion graph or learning a control policy
for the motion graph.  Using these methods, kinematic motion controllers have been created
for boxing~\cite{BoxingControllers}, and locomotion, e.g.,~\cite{Treuille, Zwicker-2008}.
Motion models may also be implicitly defined using Principle Component Analysis (PCA) of motion exemplars, 
e.g.,~\cite{Safonova}.

The space of possible motions or {\em dynamics} can also be learned as an embedding in a continuous latent space,
such as using Gaussian Process Latent Variable Models (GPLVMs)~\cite{CCC-embed} or using 
a distance metric in the original state-space~\cite{MotionFields}.  In both cases, discrete action spaces
are defined implicitly using states within a similarity neighborhood as an informal method of approximating
a distribution. Reinforcement learning for a given task can then be applied using the given motion dynamics and discrete actions.

\bfpara{\Direct, parametric models}
Parametric methods synthesize a motion, pose-by-pose, for a given behavior using a fixed-parameterization,
such as that provided by a deep neural network pose predictor; the original motion data is discarded after training.
In the direct case, the example data is considered to come from task-specific motions
and thus it can be directly used for supervised training of a sequential model for that task.
These models can maintain knowledge of the current state using explicit memory, as is the case for 
any form of recurrent neural network (RNNs), and/or direct access to the history of the sequence,
as is the case for autoregressive models. RNNs have been extensively explored for short and long-term
human motion prediction in computer vision~\cite{RNN-Frag,RNN-human}. It is commonly noted
that these models can often be unstable for long-term sequence prediction, and the production of long-term stable
sequences is considered an accomplishment, even in the absence of 
a control task~\cite{RNN-HaoLi,VAE-Holden,MT-VAE}.

For task-specific computer animation, supervised learning of (direct-predictive) parametric motion models have 
seen much recent interest.
Autoregressive DNN models can produce high-quality human variable-terrain locomotion~\cite{PFNN},
quadruped variable-terrain locomotion~\cite{MANN}, and environment aware human locomotion~\cite{NSM}.
Similarly, a mix of data augmentation and flexible objective annotation~\cite{MultiObjControl} can be used to 
learn an effective task-specific RNN model for human motion, as demonstrated on locomotion, basketball, and tennis.
\edit{
One practical consideration of direct-prediction policies is that they require careful tuning to handle run-time user requests. 
This is achieved by adding future trajectory as an input feature in~\cite{PFNN, MANN}, and object representation in~\cite{NSM}.
}

\bfpara{\Indirect, parametric models}
We can also choose to first learn a generic parametric motion model from the data, independent of the future tasks
that we may wish to use it for.  Such models support sampling from a distribution of next-state predictions,
and as such the same model can be used by a motion planner or control policy to achieve multiple desired tasks.
Mixture-density network RNNs (MDN-RNNs) output a distribution as a Gaussian mixture-model, and
 have been used as models for sequence-generation problems,
including handwriting generation~\cite{Graves-2013} and for OpenAI Gym environments~\cite{WorldModels}.
These models typically make use of autoencoders and are often autoregressive or recurrent in
the construction of the latent state.

For character animation, 
\edit{\cite{MotionGraphsPlus} uses graph traversal and probabilistic sampling techniques to synthesize motions from contact-aware Gaussian mixture motion primitives.
Our mixture-of-expert model is functionally similar but requires less predefined structure and data preprocessing.
In addition, our work also focuses on learning motion controllers in the MVAE latent space using reinforcement learning.
}
Time-convolutional autoencoders~\cite{holden2016deep} have been used to 
first learn a latent motion manifold, from task-relevant data. A mapping from 
locomotion control signals to the latent variables can then 
be trained via supervised learning. This approach does not model the forward dynamics
of the human motion, and it avoids RL by directly regressing the high-level commands 
given by the user to the learned motion features. 

Followup work develops an autoregressive and recurrent convolutional variational 
autoencoder model~\cite{VAE-Holden}. This is close in spirit to our work in many ways, 
given the similar aims and the use of conditional autoregressive VAEs. 
However, there are a number of significant differences with our work.
The stochastic latent variable is not sampled at run-time and therefore does not govern the time evolution.
The encoders and decoders have a time-convolutional structure that is absent in our work.
Control is added by direct concatenation of an encoded control signal in the latent space for use
by the decoder, and thus autoregressive modeling requires a control signal.

Other recent promising work builds a stochastic generative model for human motion using an RNN with an output distribution
modeled via the parameters of a Gaussian-Mixture Model~\cite{RNN-refine}.
The RNN output is further processed by a refiner network to remove foot skating and add robustness,
which is trained using a generative adversarial network (GAN). Control is solved as an online or offline planning problem
via initial derivative-free optimization in the sample space, followed by gradient-based optimization. 
Knowledge of contact information is assumed.  

Our MVAE model is also a parametric \indirect\ method.
We draw inspiration from the works described previously and investigate in depth how VAE-architectures
can be used for high-quality controllable real-time animation in a way that supports
reinforcement learning.
In contrast with prior work, we show that a robust stochastic generative motion model can be learned
using a memory-free first-order autoregressive model, trained with scheduled sampling. 
It generates high-quality motion without requiring contact annotations  learned post-processing, or time-convolutional structure.
\edit{
We are not aware of other parametric \indirect ~approaches using reinforcement learning that are capable of producing high-quality motions without post-processing.
}

\subsection{Physics-based motion synthesis}

In a physics-based setting, the motion model already exists, as the motion dynamics
are provided by the physics, along with a well-defined action space, often consisting of joint torques. 
A large body of recent work in deep reinforcement learning targets the learning
of control policies for physically simulated movements, either as 
motion imitation tasks~\cite{DeepLoco,DeepMimic,OnlineMimic,DReCon,BodyShape}
or without reference motion data, 
e.g.,~\cite{OpenAI-Gym,DM-parkour,BioMotion,ScalableMuscle,SymmetricLowEnergyLoco} 
and many others.
\edit{
These represent a separate stream of research and are uniquely complex in their own way.
Our work focuses on kinematics motion synthesis.
}

\section{Motion VAEs}
\label{sec:cvae-motion-model}

We develop an autoregressive conditional variational autoencoder, or Motion VAE (MVAE),
that is trained using supervised learning with motion capture data.
The MVAE implicitly models a distribution of possible next poses.  
To sample from this distribution, samples are drawn from the normally-distributed latent variables,
which are then passed through the MVAE decoder in order to realize a next-pose estimate.
Importantly, it is controllable via the choice of sample in the latent space of the learned autoencoder.
This will serve as the action space for the planner or control policy, to be described later.
The MVAE consists of an encoder and a decoder, and these two modules work cooperatively to model natural motion transitions. 
The encoder compresses high dimensional pose transition information into a compact latent representation.
The decoder takes this latent representation of the motion, as well as a condition pose, to generate the next pose.  

\bfpara{Pose Representation}  
\edit{We first compute the root position of the character by projecting the hip joint position onto the ground.
Similarly, the root facing direction is the ground projection of the forward facing axis of the hip joint.
The root position and facing direction are used to compute the character's linear and angular velocities ($\dot{r}^x, \dot{r}^y, \dot{r}^a \in \mathbb{R}$).
The joint positions ($j^p \in \mathbb{R}^3$) and velocities ($j^v \in \mathbb{R}^3$) are expressed in the character's root space.
Joint orientations ($j^r \in \mathbb{R}^6$) are represented using their forward and upward vectors in the character space.  This encoding is similar to the rotation matrix representation, which avoids the problems of angle-based representations.}
We define a pose, $p$, to be a tuple containing $(\dot{r}^x, \dot{r}^y, \dot{r}^a, j^p, j^v, j^r$).
The pose representation is similar to that defined in~\edit{\cite{MANN}}.  

\begin{figure*}%
  \centering
  \includegraphics[width=0.9\textwidth]{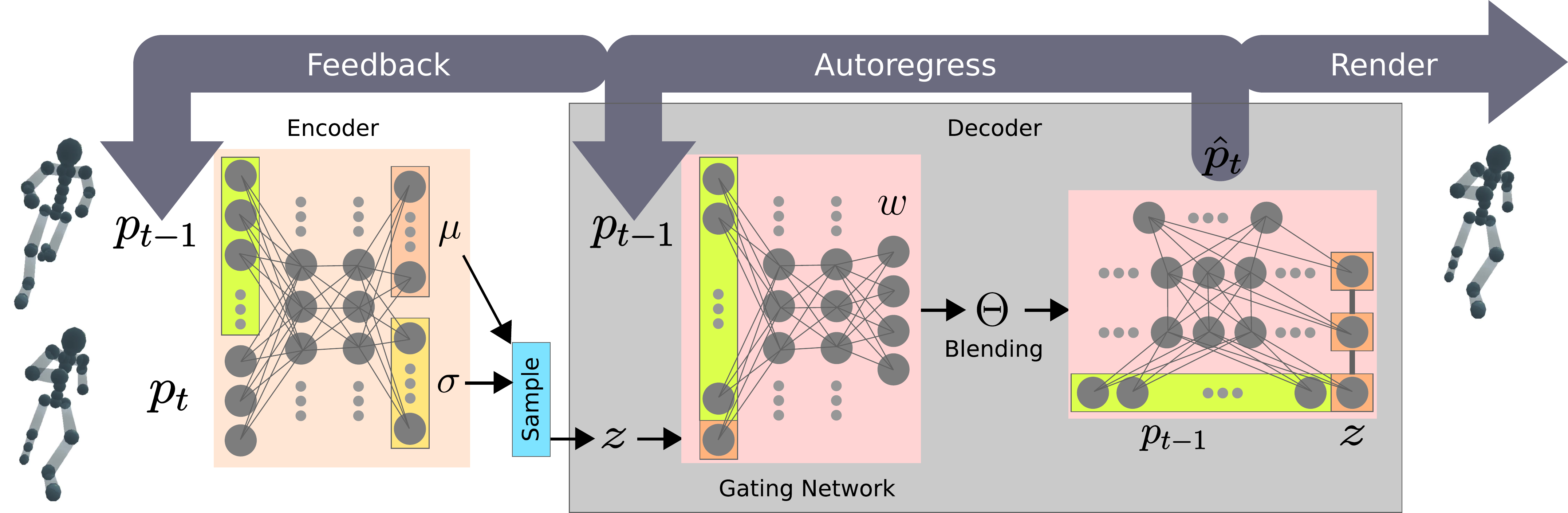}
  \caption{The conditional VAE has two parts.  The encoder takes past $(p_{t-1})$ and current $(p_{t})$ pose as input and outputs both $\mu$ and $\sigma$, which is then used to sample a latent variable $z$.  The decoder uses $p_{t-1}$ and $z$ to reconstruct $\hat{p}_t$.  For the decoder, we use a MANN-style mixture-of-expert neural network.  When using scheduled sampling during training or at run-time, the decoder output, $\hat{p}_t$, is fed back as input for generating the next prediction.}
  \label{fig:vae-model}
\end{figure*}

\bfpara{Conditions and Predictions}  
We use the MVAE to generate a distribution of possible next poses given the previous pose. 
During training, the MVAE reconstructs the next pose given the current pose, while trying to shape the latent variable $z$ into the standard normal distribution, as shown in~\autoref{fig:vae-model}.
The reconstruction loss is defined as the mean squared error (MSE) 
between the predicted pose and the next frame observed in the motion clip.
At run-time, the encoder is discarded and the decoder is used to predict future poses, one at a time, in an autoregressive fashion.
More precisely, given an initial pose, e.g.~randomly selected from the motion database, 
a sample is drawn from the latent variable $z$. This sample, together with the current pose, is used by the decoder
to generate the next pose,  which is then fed back into the decoder for the following pose, and this process repeats ad infinitum.

\subsection{Encoder Network}
The encoder is a three-layer feed-forward neural network that encodes the previous pose ($p_{t-1}$) and current pose ($p_{t}$) into a latent vector $z$.  
Each internal layer has 256 hidden units followed by ELU activations.  
The output layer has two heads, $\mu$ and $\sigma$, required for the reparameterization trick used to train variational autoencoders \cite{VAE}.  
We choose the latent dimension to be 32, \edit{which approximates the degrees of freedom of typical physics-based humanoids, such as in \cite{DeepMimic}.}
We find that the training stability and reconstructed motion quality is not overly sensitive to the encoder structure and latent dimension size.

\subsection{Decoder Network}
\label{sec:decoder}
We use a mixture-of-expert (MoE) architecture for the decoder.  
MoE methods commonly partition the problem space between a fixed number of neural network experts, 
with a gating network used to decide how much to weight the prediction of each expert when
computing a final output or prediction.
We use a style of MoE proposed in~\cite{MANN},
which we empirically observe to help achieve slightly better pose construction and reduced visual artifacts.
The MoE decoder consists of six identically structured expert networks and a single gating network to blend the weights 
of each expert to define the decoder network to be used at the current time step.
Similar to the encoder, the gating network is also a three-layer feed-forward neural network with 256 hidden units 
followed by ELU activations.  The input to the gating network is the latent variable $z$ and the previous pose $p_{t-1}$.
Each expert network is also similar to the encoder network in structure.  
These compute the current pose from the latent variable $z$, which encodes the pose transition, and the previous pose.
An important feature of the expert network is that $z$ is used as input to each layer to help prevent posterior collapse, a point we further discuss next.
\edit{Note that the gating network receives $z$ as input only for the first layer.}

\subsection{Practical Considerations}
\label{sec:cvae-considerations}

\bfpara{Avoiding Posterior Collapse}
Although we use a single pose as the condition, it is also possible to use consecutive past poses, 
i.e.~$p_{t-k}...p_{t-1}$.   In general, using multiple frames as the condition improves the 
reconstruction quality of the MVAE, but at the same time reduces the diversity of the output poses. 
In the worse case, the decoder may learn to ignore the encoder output, 
a problem known as the posterior collapse, and cause the conditional VAE to only playback the original motion capture data. 
We find that using one or two consecutive poses as the condition works well for our experiments, 
but in general, the optimal choice may be a function of diversity and quality of the motion database.
To further prevent posterior collapse, we emphasize the importance of the latent variable by 
passing it to every layer of the \edit{expert} network, as described in~\edit{\autoref{sec:decoder}}.
We find empirically that this trick reduces the likelihood of posterior collapse happening.
Other design decisions that impact the likelihood of posterior collapse include the decoder network size, 
the weight of the KL-divergence loss (in $\beta$-VAE), and the number of latent dimensions.

\bfpara{Balancing Motion Quality and Generalization}
A fundamental challenge of all kinematic animation systems is the need to balance motion quality against generalization.  
In a non-parametric setting, e.g.,~\cite{MotionGraphs,MotionFields,MotionMatching},
this can be adjusted by tuning the number of nearest-neighbors and the distance threshold in the nearest neighbor search.  
In VAEs, the balance comes from weighting the reconstruction  and  KL-divergence losses.  
When motion quality is heavily favored, the system simply replays the original motion capture sequences, 
and as a result, it will not respond effectively to user control.  
Conversely, when motion generalization is favored, the system may produce implausible poses and motions.  
A generative model that can generalize to all physically-feasible motions must learn 
to infer the laws of Newtonian mechanics, which is difficult given limited training data.
Therefore, when data is limited, the goal is to strike a balance between motion quality and generalization.  
We find that having the MVAE reconstruction and KL-divergence losses be within 
one order of magnitude of each other at convergence is a good proxy for finding an appropriate
balance between quality and generalization.

\subsection{MVAE Training}
\label{sec:cvae-training}

\begin{figure}%
  \centering
  \includegraphics[width=1\columnwidth]{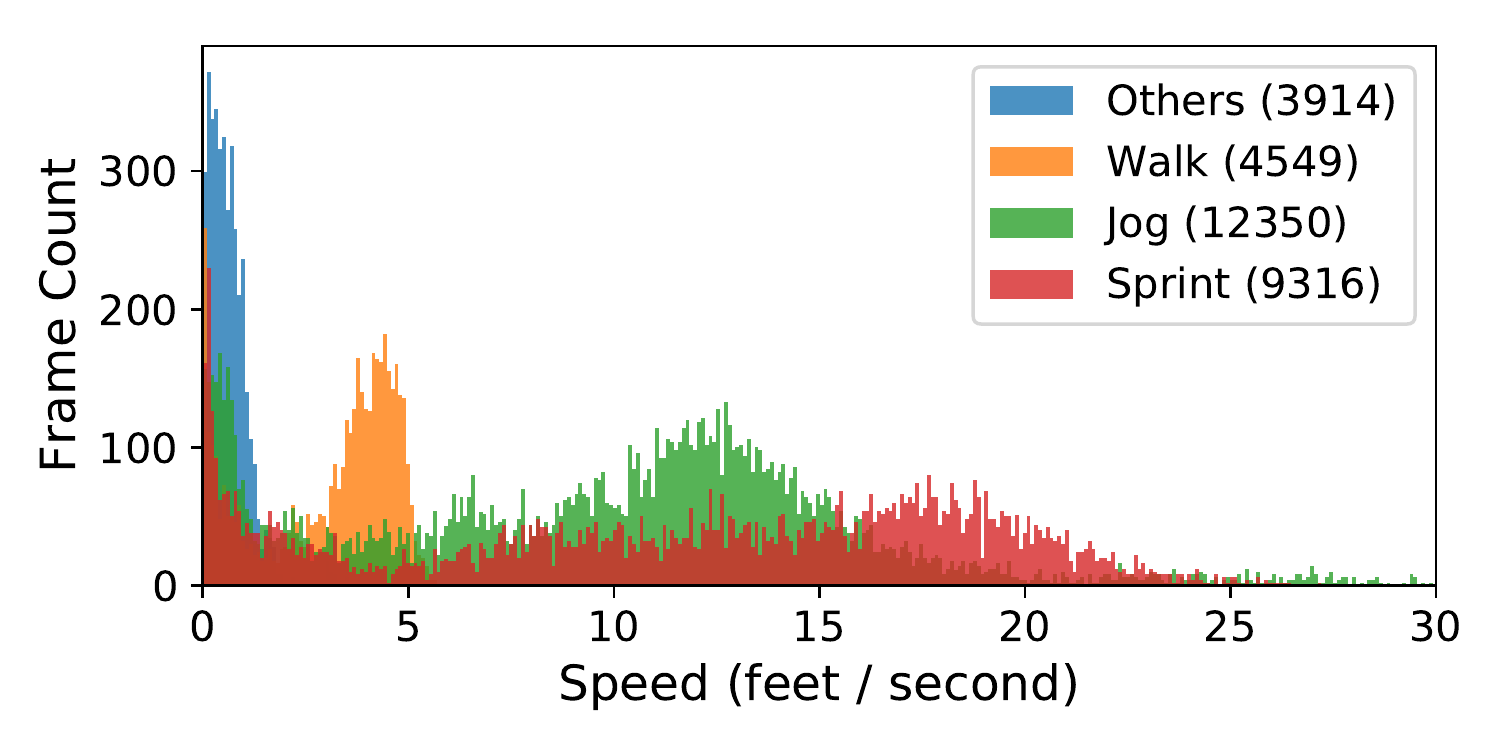}
  \caption{A breakdown of different motions used for training.  The number of frames in each category is labeled in brackets.  The \textit{Others} category contains mostly non-locomotion motion clips, such as in-place turning and resting.}
  \label{fig:mocap-distribution}
\end{figure}

Our motion capture database contains 17 minutes of walking, running, turning, dynamic stopping, and resting motions.
This includes the mirrored version of each trajectory.
The data is captured at 30~Hz and contains about 30,000 frames.
The motion classification, i.e.~walking and running, is not used during training and 
there is no further preprocessing, i.e.~foot contact and gait-phase annotations are not required.
The breakdown of the motion capture database clips is visualized in \autoref{fig:mocap-distribution}.

The training procedure follows that of a standard $\beta$-VAE.  
The objective is to minimize the reconstruction and KL-divergence losses.  
We choose $\beta=0.2$ to minimize the chance of posterior collapse.
We note that the learning stability is not sensitive to the exact value of $\beta$.
We find that better generalization occurs when z-score normalization is applied to the training data.
Intuitively, standardizing the input data reduces the bias caused by the motion range differences of each joint.
This is analogous to the pose similarity metrics used in~\cite{MotionFields}, which uses
independent scaling factors for each joint in proportion to the bone length.
We use Adam optimizer \cite{Adam} to update the network weights.  
The learning rate is initialized at $10^{-4}$ and is linearly decayed to zero over 180 epochs.  
With a mini-batch size of 64, the entire training procedure takes roughly two hours on an 
Nvidia GeForce GTX 1060 and an Intel i7-5960X CPU.

\begin{figure}%
  \centering
  \includegraphics[width=1\columnwidth]{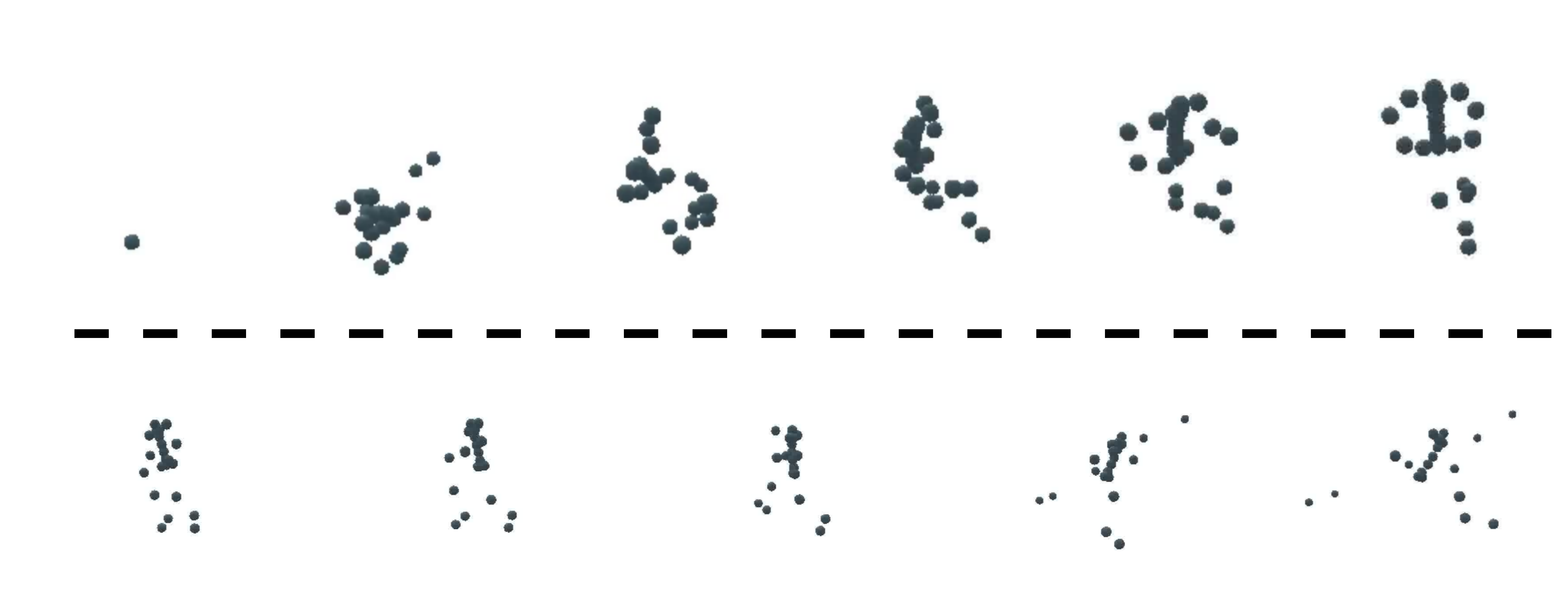}
  \caption{Comparison of MVAE reconstruction stability with and without scheduled sampling.  The time axis flows from left to right.  \textbf{Top}: With scheduled sampling, the MVAE is able to reconstruct the character body even when the initial joint locations are all clustered into a single point.  \textbf{Bottom}: Pose reconstruction of MVAE trained without scheduled sampling with exactly the same settings for the same number of epochs.  Even though the character is initialized to a motion capture frame, reconstruction error still quickly accumulates causing the joints to fly apart.}
  \label{fig:vae-reconstruction}
\end{figure}

\bfpara{Stable Sequence Prediction}
The MVAE trained with standard supervised learning suffers from unstable predictions when 
making autoregressive predictions at run-time.  This is due to 
growing reconstruction errors that can rapidly cause the MVAE to enter a new and unrecoverable region of the state space.
The consequence is clear when visualizing the predicted poses, which no longer resemble a character body. 
To alleviate this, we use {\it scheduled sampling} \cite{SchedSamp} to progressively introduce 
the run-time distribution during training.  
A sample probability $p$ is defined for each training epoch.  
After a pose prediction is made, it is used as input for the next time step with probability $1-p$,
instead of using the pose from the ground truth training data. 
The entire training process is divided into three modes: supervised learning ($p=1$), scheduled sampling (decaying $p$), and autoregressive prediction ($p=0$).  
The number of epochs for each mode is 20, 20, and 140 respectively.  
For the scheduled sampling mode, the sampling probability decays to zero in a linear fashion with each learning iteration.
\autoref{fig:vae-reconstruction} illustrates the ability of the MVAE to recover after being
trained in this fashion.

\bfpara{Training MVAE with Mini-batches}
In order to implement scheduled sampling, we need to define a prediction length $L$ for each roll-out, 
even though we do not use a recurrent neural network.
In each roll-out, we randomly sample a start frame from the motion capture database and perform pose prediction for $L$ steps.
Ideally, $L$ should be large enough to allow prediction errors to accumulate, so as to simulate the actual run-time distribution.
We find that training with $L=8$ \edit{($1/4$ second)} is enough to prevent covariate shift for our motions.  
Another technical detail for doing mini-batched supervised learning is that we need to handle the end-of-clip problem.
Since we know in advance that each frame sample needs $L=8$ subsequent frames, 
we can choose to only sample frames that meet this condition.
A common practice in sequential prediction supervised learning tasks
is to cope with variable length inputs by padding the training sequences to match the length of the longest sequence,
e.g., for sentence generation in natural language processing, 
where the end of input sentences are padded with end-of-sentence tokens.  
In our case, padding would be inappropriate because, 
unlike sentences, the motion clips we used do not have a definitive end pose.  
Since any frame can be treated as a start and end pose of a sequence, 
the motion clips can be freely divided into equal-length sequences.
However, this assumption may not hold for some motions where the start and end poses need to be fixed, such as gymnastic movements.

\section{Motion Synthesis}
\label{sec:motion-synthesis}

With the learned MVAE motion model in place, we turn to the problem of control.
As noted previously, this can be achieved via the space of possible actions,
as defined by samples from the latent encoded state $z$.
In this section, we present two simple control strategies, random sampling and sampling-based control. 
In the following section, we then present the use of
reinforcement learning methods on top of the MVAE model.
At this point, the MVAE network weights are fixed. 
Only the decoder is used to generate pose sequences; the encoder is discarded.

\subsection{Random Walk}

\begin{figure}%
  \centering
  \includegraphics[width=1.0\columnwidth]{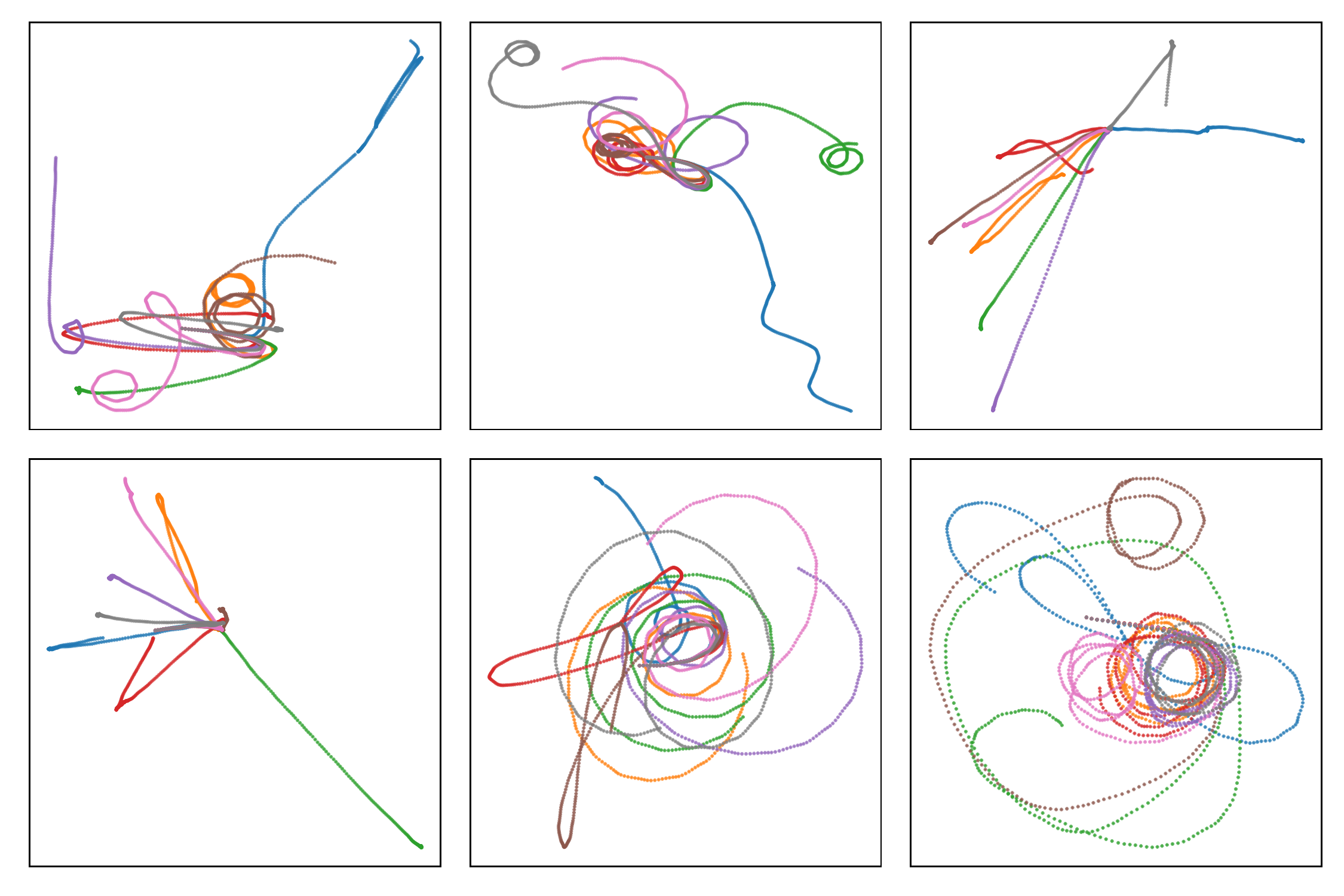}
  \caption{Random walks visualized for six different initial conditions. Each of eight characters
is initialized to the same initial condition, but unique latent variable samples are drawn at each time step for 300 steps.  
The resulting root trajectories are shown, with the original motion capture trajectory shown in blue for comparison.}
  \label{fig:random-walk-root}
\end{figure}

Given an initial character state, we can simulate random plausible movements
by using random samples from the MVAE latent distribution.  
Even with our single frame condition, the synthesized reconstructed motion will typically resemble 
that of the original motion clip from which the starting frame is chosen.
E.g., when the initial character state comes from the middle of a sprint cycle, 
the MVAE will continue to reconstruct the sprint cycle.
Furthermore, when the initial state is a stationary pose -- a common pose at the start of most motion clips -- 
the character can transition into walking, running, jumping, and resting motions.
\autoref{fig:random-walk-root} shows the effect of the conditioning on the 
reconstructed motion.  Examples are also shown in the supplementary video.

One challenge in kinematic animation is to 
know whether the underlying motion database is capable of a given desired motion.
For instance, it may not obvious when a data-driven animation system
can transition between two motion clips.
Often we might believe that two motion clips have close enough transition points upon visual inspection, 
but the actual distance in high-dimensional pose space may not match our intuition.
In the random walk condition, it can be plainly observed when a particular motion is isolated in the pose space 
and therefore has no transition to other motions. 
When the random walk is unable to transition despite drawing many samples, 
it is an indication that additional motion capture data, especially the transition motions, may need to be supplied.
An advantage of using an MVAE model is that it is small enough, and therefore fast enough, for quick design iterations.
In our experiment, we used this method to find that our original motion capture database 
had an insufficient number of character turning examples.

\subsection{Sampling-based Control}

\renewcommand{\sectionautorefname}{\S}
\let\subsectionautorefname\sectionautorefname 
\let\subsubsectionautorefname\sectionautorefname

We next develop a simple sampling-based controller. This 
performs multiple Monte Carlo roll-outs ($N$) for a fixed horizon ($H$).
\edit{The first action of the best trajectory, among all sampled roll-outs,} is selected and applied to the character for the current time step.
This procedure is then repeated until the task is accomplished or terminated.
We find this simple sampling-based control method works modestly well for simple locomotion tasks, such as Target (\autoref{sec:env-target}).
Using $N=200$ and $H=4$, the character can generally navigate towards and circle around the target, 
as well as adapting to sudden changes in the target location. This is shown in the supplementary video.

When compared to policies learned with RL (\autoref{sec:character-control}), the policy has difficulty directing the character to reach within two feet of the target.
For more difficult tasks, such as Joystick Control (\autoref{sec:env-joystick-control}) and Path Follower (\autoref{sec:env-path-follower}), the simple sampling-based policy is unable to achieve the desired goals.
In such scenarios, a more sophisticated approach, such as \cite{rajamaki2017augmenting}, would likely be able to find better solutions.
In general, fine-tuned sampling-based methods can provide faster design iteration cycles for artists at the cost of more run-time computation. 

\renewcommand{\sectionautorefname}{Section}
\let\subsectionautorefname\sectionautorefname 
\let\subsubsectionautorefname\sectionautorefname

\section{Learning Control Policies}
\label{sec:character-control}

\begin{figure}
  \centering
  \includegraphics[width=0.95\columnwidth]{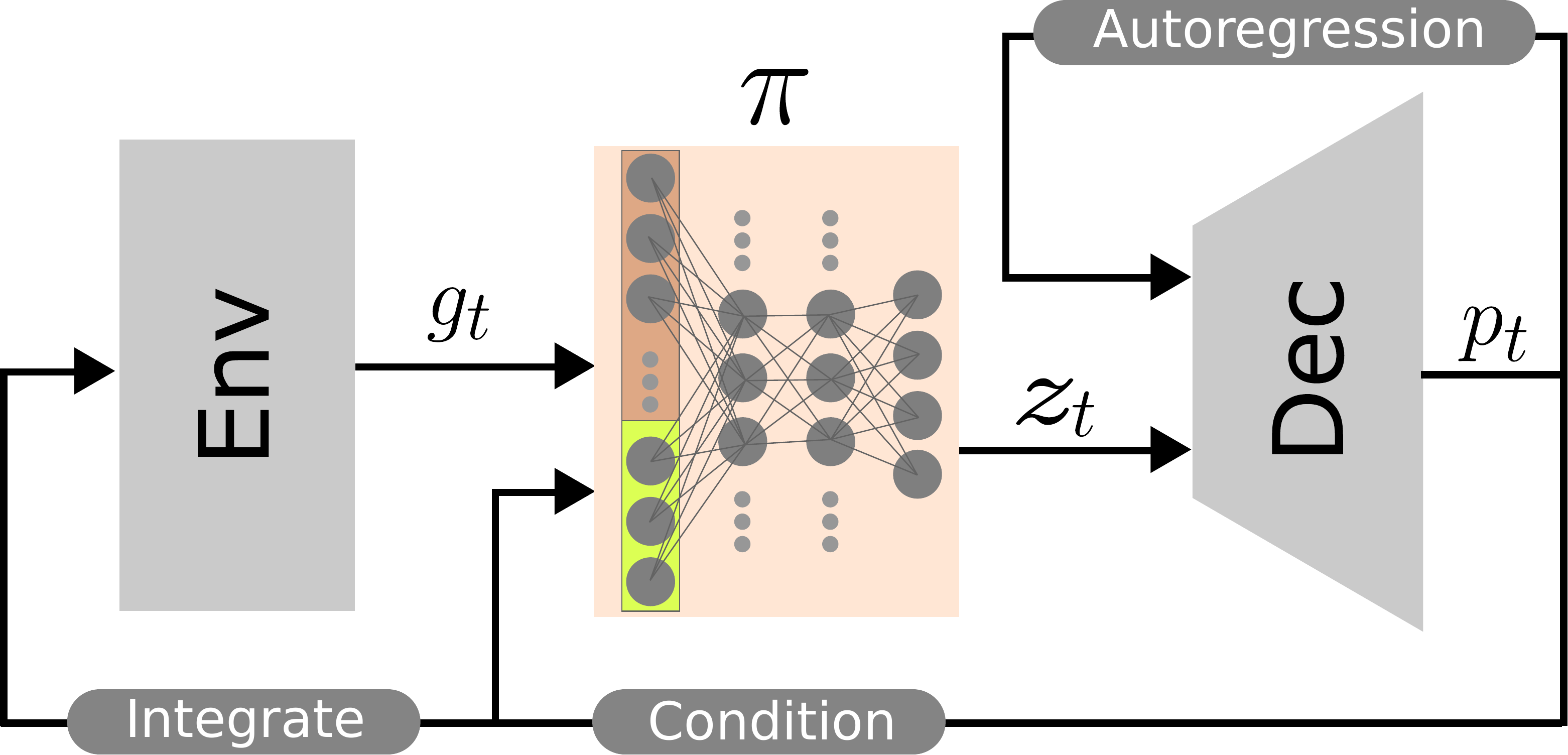}
  \caption{Using DRL, we can learn a policy to compute the latent variable $z$ at every step to guide the character.  The encoder is discarded since we do not have the next pose from the motion capture sequence.  Initially, a random pose $p_0$ is selected from the motion capture database.  In each time step, the environment uses $p_t$ generated by the decoder to integrate the root positions and computes relevant task information $g_t$.  The policy $\pi$ computes the action from $p_t$ and $g_t$, which is fed as $z_t$ back into the decoder.}
  \label{fig:vae-controller}
\end{figure}

A control policy can be used to guide the character to perform various tasks
using the latent samples $z$ as an action space. 
This is illustrated in \autoref{fig:vae-controller}.
Unlike direct-prediction policies, e.g.,~\cite{PFNN,MultiObjControl}, a control policy
can be learned in support of arbitrarily-defined rewards that can be used to shape a behavior.
\renewcommand{\sectionautorefname}{\S}
\let\subsectionautorefname\sectionautorefname 
\let\subsubsectionautorefname\sectionautorefname
\edit{We demonstrate the flexibility of our system in handling a variety of task representations (\autoref{sec:tasks}), e.g.~target location, desired speed and direction, and working with local-vision sensing capability.}
\renewcommand{\sectionautorefname}{Section}
\let\subsectionautorefname\sectionautorefname 
\let\subsubsectionautorefname\sectionautorefname

\paragraph{Reinforcement Learning}
We use deep reinforcement learning (DRL) to learn various locomotion skills. 
In RL, at each time step $t$, the agent reacts to an environment state $s_t$ by performing an action $a_t$.
Based on the action performed, the agent receives a reward signal $r_t = r(s_t, a_t)$ as feedback. 
In DRL, the agent computes $a_t$ using a neural network policy $\pi_\theta(a | s)$, where $\pi_\theta(a | \cdot)$ is the probability density of $a$ under the current policy. 
The goal of DRL is to find the network parameters $\theta$ which maximize the following:

\begin{align*}
J_{RL}(\theta) = E\left[\sum_{t=0}^{\infty}\gamma^t{r({s}_t, {a}_t)} \right].\\
\end{align*}

Here, $\gamma \in [0, 1)$ is the discount factor so that the sum converges.
We solve this optimization problem using the proximal policy optimization (PPO) algorithm~\cite{PPO}.
We use a publicly available implementation of PPO \edit{\cite{pytorchrl}}.

\subsection{Controller Network}
The control policy is a two hidden-layer neural network with 256 hidden units followed by ReLU activations.
The output layer is normalized with Tanh activation and then scaled to be between -4 and +4.
Since the policy outputs the latent variable, the scaling factor should be chosen according to the latent distribution.
In the case of a CVAE, the latent distribution is assumed to be a normal distribution with zero mean and unit variance.
With a scaling factor of 4, the majority of the probability mass of a standard normal distribution is covered.
The value function network is identical to the policy network, except for the output layer width and normalization.

The convergence of DRL algorithms can be sensitive to hyperparameter settings in general, 
but we do not find that to be the case in our experiments.
We use the same hyperparameters for every task.
The learning rate decays exponentially following the equation: $\text{max}(1, 3 \cdot 0.99 ^ \text{iteration}) \times 10^{-5}$.
We find that decaying the learning rate exponentially helps to improve the training stability of the policy network.
For collecting training data, we run 100 parallel simulations until the episode terminates upon 
reaching the maximum time step defined for each task, typically 1000.
The policy and value function networks are updated in mini-batches of 1000 samples.
Since all computations are done on the GPU, the data collection and training processes are fast despite the large batch size.
All tasks described in the following sections can be fully trained within one to six hours on our desktop machine.

\subsection{Effort Penalty in Kinematics Animation} 
In physics-based animation control, an energy or effort penalty is often used to restrict the solution space 
such that the learned policy produces natural motions.
Mechanical energy can be easily calculated from torque, since torque is already used as part of the physics simulation.
In contrast, it can be difficult to define an energy term in kinematic animation, so typically root velocity is used as a proxy.
In our motion data, we find that the root velocity metrics can often be inconsistent with our intuition 
of physical effort when visually examining the motion.
To accurately quantify effort, we should consider the motion of the joints as well, and we therefore define energy as follows:

\begin{align*}
E = (\dot{r}^x)^2 + (\dot{r}^y)^2 + (\dot{r}^a)^2 + \frac{1}{J} \sum_j^J \norm{j^v}^2.\\
\end{align*}

A more accurate energy metric should take masses and inertia of each joint into consideration.
However, we find that scaling the individual contributions of the joint energy terms is approximately equivalent.
When we include the energy measure as a penalty in RL optimization, we see that the policy is able 
to find visibly lower effort solutions.
In comparison to the common approach of using a target root velocity to regulate character energy expenditure, 
our approach of using RL to optimize for energy is more natural.
Also, since the target velocity does not need to be supplied by the user at run-time, 
our approach is more flexible for animating non-directly controllable 
characters or large crowds. Please refer to the supplementary video for 
the impact of energy-based regularization.

\section{Locomotion Controllers}
\label{sec:tasks}
We now describe multiple locomotion tasks that can be achieved using learned RL-based control policies
on top of the MVAE model.
\renewcommand{\sectionautorefname}{\S}
\let\subsectionautorefname\sectionautorefname 
\let\subsubsectionautorefname\sectionautorefname
The locomotion tasks are: Target (\autoref{sec:env-target}), Joystick Control (\autoref{sec:env-joystick-control}), Path Follower (\autoref{sec:env-path-follower}), and Maze Runner (\autoref{sec:env-maze-runner}).

\renewcommand{\sectionautorefname}{Section}
\let\subsectionautorefname\sectionautorefname 
\let\subsubsectionautorefname\sectionautorefname

\subsection{Target}
\label{sec:env-target}

The goal for this task is to navigate towards a target that is randomly placed within the bounds of a predefined arena.
The character begins in the center of the arena and knows the precise location of the target at any given moment in its root space.
Upon reaching the target, a new target location is randomly selected and the cycle starts over again.
We define the character as having reached the target if its pelvis is within two feet of the target.
Furthermore, we define the size of the arena to be $120 \times 80$ feet to simulate the fact that soccer pitches are rectangular.
The exact values of these parameters do not impact learning and solving of the task.

In the context of \autoref{fig:vae-controller}, the environment needs to compute the goal $g_t$ and the reward $r_t$ in each time step.
The environment keeps track of the root position and root facing of the character, i.e.~$r^x$, $r^y$, and $r^a$, in global space.
In each time step, the environment first computes the new global state of the character by integrating $\dot{r}^x$, $\dot{r}^y$, and $\dot{r}^a$ in the current pose.
The coordinate of the target in character root space is provided to the policy.
The reward, $r(s, a)$, is a combination of progress made towards the target and a bonus reward for reaching the target.
The progress term is computed as the change in distance to the target in each step after applying the integration procedure.
Upon reaching the target, the policy receives an one-time bonus before a new target is set.

\bfpara{Visualizing the Value Function}

\begin{figure}%
  \centering
  \includegraphics[width=1.0\columnwidth]{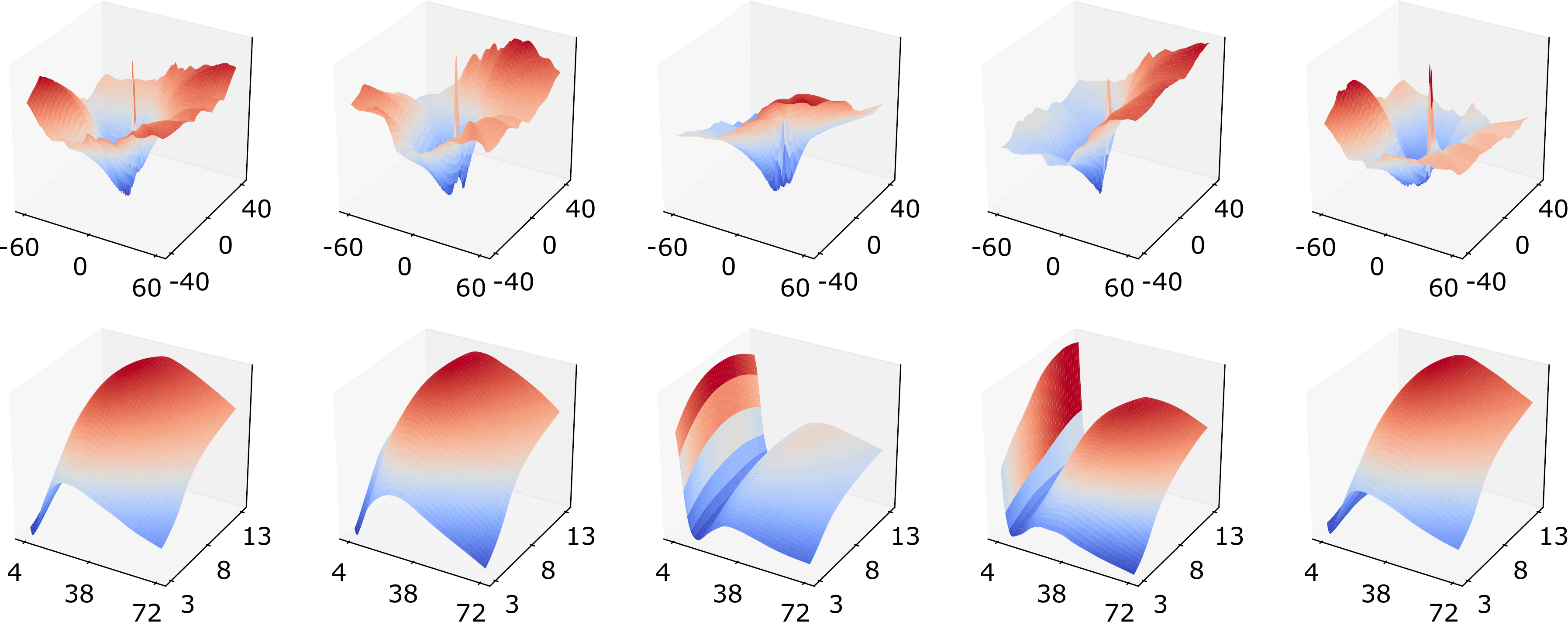}
  \caption{Visualizing the value function for the Target task.  \textbf{Top}: Using a surface plot, we can show the value function computed if the target is placed at coordinates $(x, y)$, corresponding to the left and right axis.  Each surface plot shows a different character initialization -- From left to right, we have forward sprint, 180\textdegree~turn, right banking turn, left turn, and resting.  \textbf{Bottom}: Value functions for additional timed-variant experiment of the Target task in which the character has limited time to reach the targets.  The left axis corresponds to the target distance in feet and right axis to the remaining time in seconds.} 
  \label{fig:value-function}
\end{figure}

In actor-critic based RL algorithms, a value function is learned alongside the policy.
Although at test time we only need to query the policy to get the action, 
the value function contains valuable information regarding the limitations of the policy.
\autoref{fig:value-function} shows the value function for target locations sampled across 
the arena and different character initial states.
The character is always located at the origin and facing along the x-axis.
We see that the value function peaks at the origin as expected;
if the target is also located at the origin, then the character does not need to move at all to receive the target bonus.
Moreover, we see that the value function is lower in the vicinity of the character and rises gradually as the target gets further.
This means that it is easier and quicker for the policy to navigate the character to targets that are at a certain distance.
Considering the data distribution of the original motion capture sequences 
which mostly contain running motions, this observation is reasonable.
Lastly, we also observe reasonable value function shapes when the character is initialized to a left or right turn.

We further experimented with a timed-variant of the Target task in which the character has limited time to reach the targets.
Each of the value function plots in \autoref{fig:value-function} contains an inverted parabolic cylinder-like surface, e.g.~the expected return is highest at middle distances.
Again, this is the combined effect of the progress reward and target bonus.
For a fixed time slice, the expected return is low when the target is close because the total receivable progress reward, i.e.~the potential, is small.
Conversely, when the target is beyond reach given the remaining time, the contribution from the target bonus vanishes from the value function.

\bfpara{Testing MVAE Generalization}
As discussed in \autoref{sec:cvae-considerations}, a trained MVAE model may not always 
generalize enough to allow the character movement to be controlled.
While this is clear when posterior collapse occurs, it can be less obvious if the MVAE still exhibits some generalization.
We devise a straight line running task as a sanity check to determine if the model has enough generalization for learning a controller.
This task is a special case of the Target task where the character is initialized to be at the start of a running sequence and the target is placed directly in front of it.
By varying the distance between the character and the target, we can test whether the MVAE has sufficient controllability.
Please see the supplementary video for more detail.

\subsection{Joystick Control}
\label{sec:env-joystick-control}

Joystick control is another standard locomotion task in animation.  
The task requires the character to change its heading direction to match the direction of the joystick and adjust its forward speed proportional to the magnitude of the joystick tilt.
Note that this is, in essence, the default task in previous work for bipeds~\cite{PFNN} and 
quadruped characters~\cite{MANN}.
In those works, a future trajectory can be generated from the joystick and character state at test time.
The ability to use RL means that this desired task can be defined more directly.

We simulate joystick control by changing the desired direction and speed every 120 and 240 frames respectively.
The desired direction ($r^a_d$) is uniformly sampled between 0 and $2\pi$, regardless of the current facing direction.
For the desired speed ($\dot{r}_d$), we uniformly select a value between 0 and 24 feet per second, which is the typical velocity range for our character in the example motions.
The character receives the following reward in every time step,
\begin{align*}
r_{joystick} = e^{\text{cos}(r^a - r^a_d) - 1} \times e^{-|\dot{r} - \dot{r}_d|},
\end{align*}
where $\dot{r}$ is the current speed of the character. 
The cosine operator in the first reward term addresses the discontinuity at 0 and $2\pi$ when calculating difference between two angles, while the two exponentials are used to normalize the reward to be between 0 and 1.
Multiplying the two reward terms encourages the policy to satisfy both target direction 
and speed simultaneously~\cite{ScalableMuscle}.
At run-time, the user can control the desired heading direction and speed interactively.

\subsection{Path Follower}
\label{sec:env-path-follower}

In the Path Follower task, the character is required to follow a predefined 2D path as closely as possible.
We implement this task as an extension of the Target task, 
with the character seeing multiple targets ($N=4$), each spaced 15 time steps apart, along the predefined path.
We feed the policy $N$ target locations, rather than the entire path, so that it does not memorize the entire trajectory.
This way the policy can have a chance to adapt to path variations at run-time without further training.

\begin{figure}%
  \centering
  \includegraphics[width=1\columnwidth]{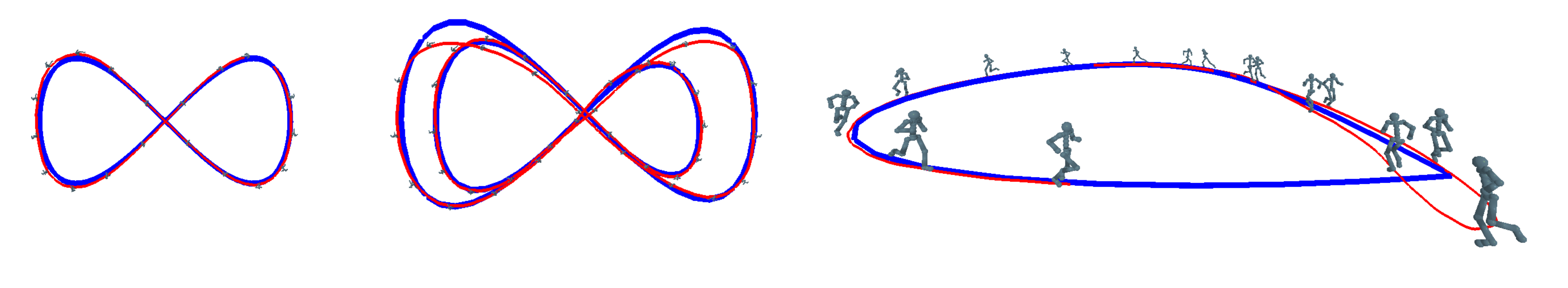}
  \caption{Character performing the Path Follower task.  \textbf{Left}: Top view of the character root trajectory (red) overlaid on top of a \textit{figure 8} path (blue).  Overall the character is able to match the curve well and remain on target.  \textbf{Middle}: The policy is able to adapt to new paths with different curvatures and speed requirements. Since the goal for the character is to chase a moving target, the policy exhibits corner-cutting behaviour when the target is moving faster beyond the capability of the character.  \textbf{Right}: The policy is unable to find an 180-degree turn, so a small radius looping motion is improvised to keep the character close to the moving target.}
  \label{fig:path-follower}
\end{figure}

We train the policy on a parametric \textit{figure 8}, given by $x = A\,\text{sin}(bt)$ and $y = A\,\text{sin}(bt)\,\text{cos}(bt)$ where $t \in [0, 2\pi]$, $A = 50$, and $b = 2$.
The time step is discretized into 1200 equal steps.  
We choose this particular curve because it contains left and right turns, as well as straight line segments that require the character to adjust its speed in and out of the turns.
It is important to note that the targets advance with time, regardless of the current location of the character.
Therefore, the policy must learn to speed up and slow down to match the progression of the curve, as well as learn to recover if it has deviated from the path.
We find randomizing the initial position of the character to be important, in a way that is analogous to reference state initialization~\cite{DeepMimic}.
In addition, we set the initial orientation of the character to match the direction of the curve.
In the absence of this, the policy may never learn the later segments of the path.
\autoref{fig:path-follower} shows that the character can generally stick to the path except for a few challenging scenarios.

\subsection{Maze Runner}
\label{sec:env-maze-runner}

All previous locomotion tasks have explicit and structured goals in the form of target location and desired direction and velocity.
In contrast, the Maze Runner task allows the character to freely explore the space within the confines of a predefined maze.
Different from traditional RL maze environments, such as AntMaze~\cite{frans2017meta} and others~\cite{GoExplore}, our maze is fully enclosed without an entrance or exit.
The character begins at a random location in the maze and is rewarded for covering as much area as possible.

\begin{figure}
  \centering
  \includegraphics[width=0.95\columnwidth]{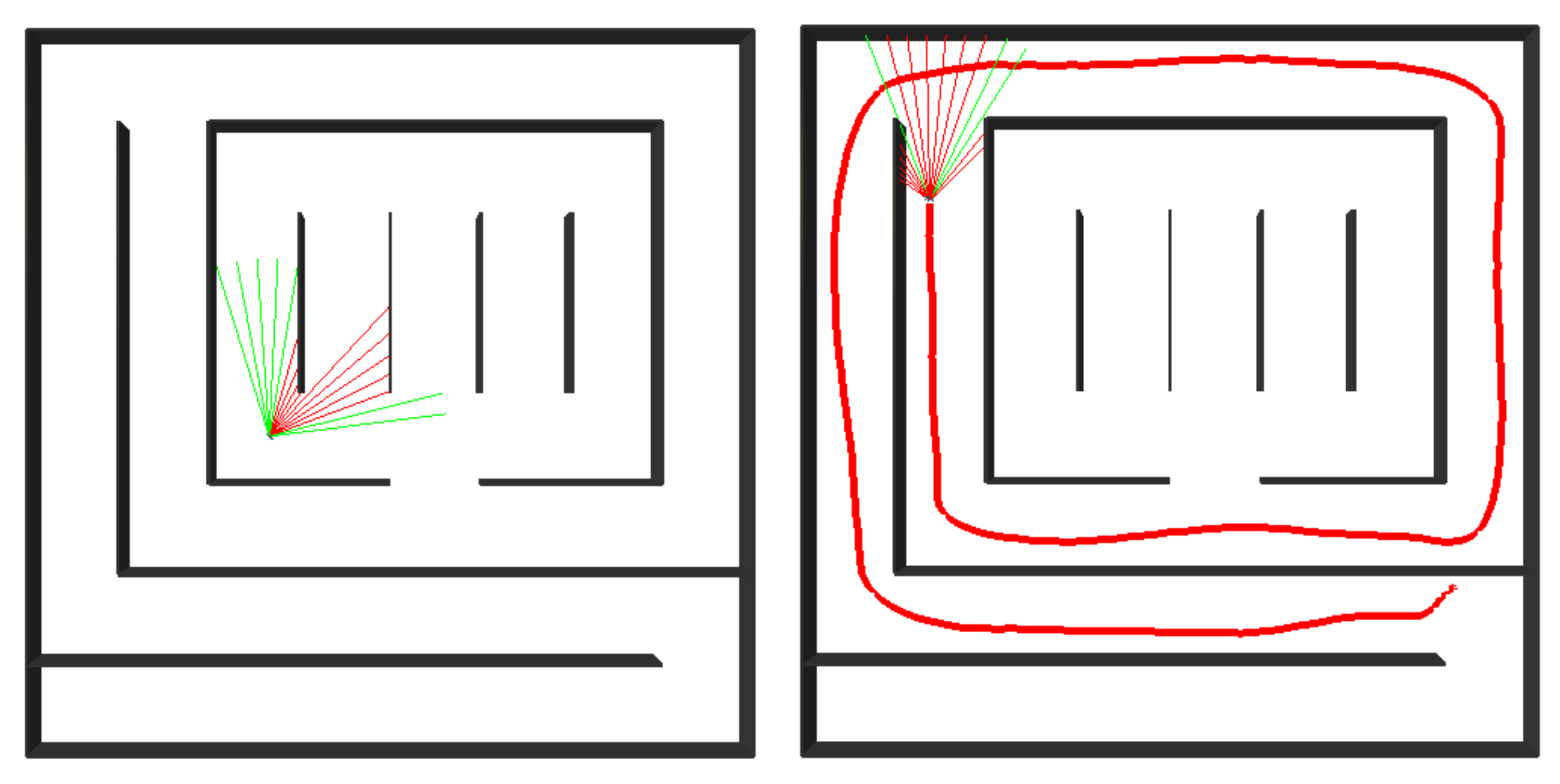}
  \caption{The Maze Runner environment requires the character to explore within the maze as much as possible.  \textbf{Left}: The vision system includes 16 rays; each encodes a distance to the nearest wall, up to a maximum distance of 50 feet. \textbf{Right}: The character can be initialized anywhere within the maze. The red path traces the progress made by the character thus far in the current episode.}
  \label{fig:maze-runner}
\end{figure}

The arena is a square of $160 \times 160$ feet and the total allotted time is 1500 steps.
For simplicity, we define an exploration reward by dividing the maze into $32 \times 32$ equal sectors.
Each time the character enters a new sector, it receives a small bonus.
The exploration reward can be viewed as a bias for encouraging the policy to be in constant motion, without explicitly specifying how the character should move.
The task is terminated immediately when the character hits any of the walls, or when the allotted time is exhausted.
Rather than receiving explicit target locations, the character uses a simple vision system to navigate in the environment.
The vision system involves casting 16 light rays centred around the current facing direction.
Each light ray is encoded by a single floating-point number, which represents the distance to the wall, up to a maximum distance of 50 feet.
Note that the character is unaware of its current position in the global space, therefore it must rely on its simple vision 
system to avoid obstacles.
\autoref{fig:maze-runner} shows that, with the exploration reward, the policy learns a non-stationary solution and is capable of avoiding walls using the vision system.

We find hierarchical RL to be beneficial for solving this task. 
Without it, the policy often fails to avoid colliding with the walls even at convergence.
To this end, we train a high-level controller (HLC) on top of a pre-trained low-level controller (LLC) for the Target task, similar to \cite{DeepLoco}.
The HLC outputs a target location at each time step, which is consumed by the LLC to compute an action.
Since both HLC and LLC operate at the same control frequency, this suggests that the hierarchical approach may be 
unnecessary given better fine-tuning of a single policy network.

\section{Discussion \protect\edit{\& Evaluation}}

This section presents observations regarding the capabilities of the learned MVAE model 
when combined with reinforcement learning to produce better quality animations.

\bfpara{Why not use an RNN?\nopunct}
Recurrent neural networks (RNN), such as LSTMs and GRUs, are a standard learning-based approach to working with sequential data.
While RNNs have seen success in areas such as NLP, audio, and video, 
our work demonstrates that a first-order autoregressive model can be sufficient for working with sequential pose data.
An advantage of RNNs is that they can encode sequence information from the past as hidden states 
and use the information in subsequent predictions.
However, this is unnecessary in fully observable dynamic simulations. 
The motion of rigid bodies follows the Markov assumption where the next state depends only on the current state and action.
Although the Markov condition may not hold for kinematic animation, 
the pose vector may already contain enough information such that the benefit of having a hidden state is minimal.
Many RNN-based motion synthesis methods produce deterministic single-pose estimates, and not the
distributions needed for modeling the space of available motion transitions.

\begin{figure}
  \centering
  \includegraphics[width=1\columnwidth]{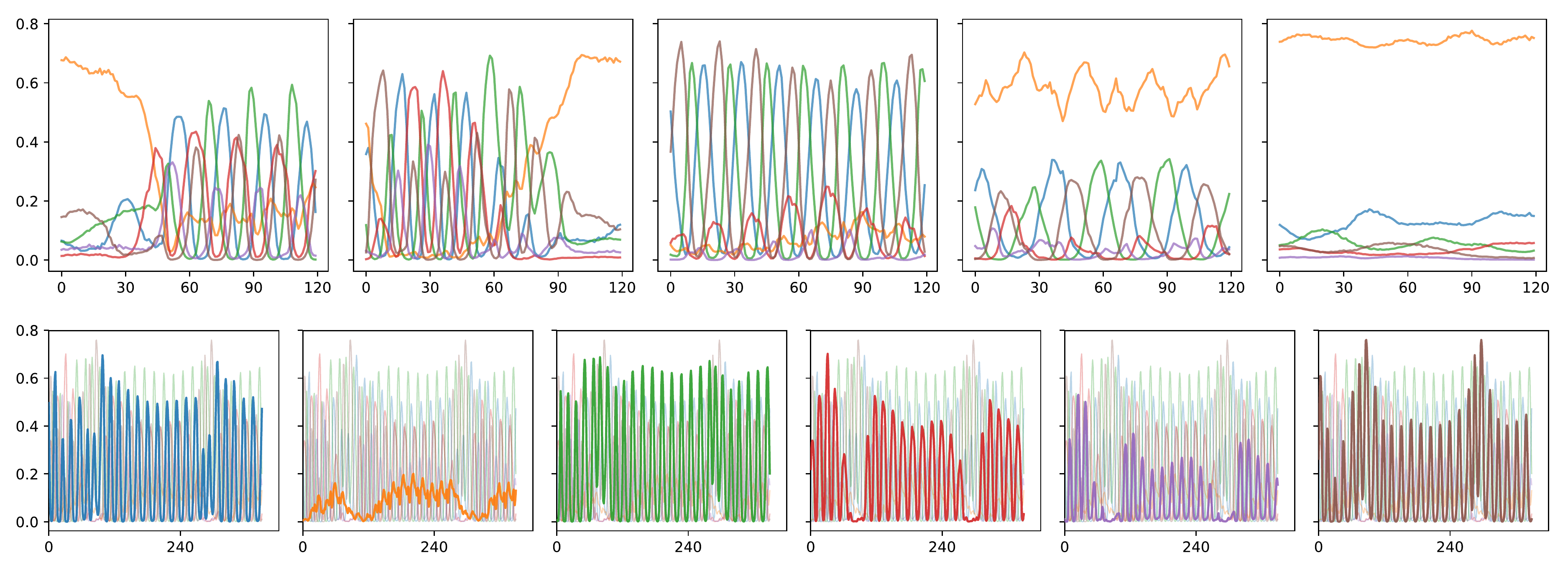}
  \caption{\textbf{Top}: Visualization of blending coefficients under random walk starting from various initial poses.  The initial poses, from left to right, as the same as the ones in \autoref{fig:value-function}.  \textbf{Bottom}: Visualizing expert utility under policy control in Target task.  From left to right, each figure highlights the blending coefficient of the corresponding expert.  Please see the supplementary video for further detail.}
  \label{fig:gating-pattern}
\end{figure}

\bfpara{Patterns in the Gating Network}
\autoref{fig:gating-pattern} shows that the blending coefficients exhibit clear sinusoidal patterns and that each expert is 
utilized approximately equally.
In the random walk scenario, we visualize the blending coefficients when the character is performing different actions, i.e.~sprinting, 180\textdegree~turn, right banking turn, left turn, and resting.
In the sprinting motion, we can see a clear transition from the preparation phase into a run cycle, 
even just by looking at the expert utility plot.
For the resting motion, the non-periodic nature of the curves is consistent with the acyclic nature of the motion.

Another pattern emerges when we visualize expert activation under policy control when solving the Target task.
We plot the blending coefficients on a larger time scale in the second set of figures in \autoref{fig:gating-pattern}.
Each expert activation curve resembles a high-frequency sinusoidal signal enveloped by a low-frequency signal.
We find the peaks and troughs of the high-frequency oscillation to be consistent with the foot strike timing.
Furthermore, the low-frequency envelope corresponds to the overall character motion.
In particular, the troughs on the orange curve (i.e.~expert 2) and the peaks on the green curve are consistent with the character performing a turn after reaching the target.
The emergence of these structures means that the MVAE not only learned to reconstruct the character pose, 
but was also able to infer locomotion features such as foot contact.
This observation opens a promising approach to learn character motion. 
If during training of the MVAE and the policy we can manipulate the underlying signal, 
such as making the transitions sharper or the period longer, 
then we may be able to achieve more efficient learning and higher quality motions.

\subsection{System Evaluation}
\label{sec:evaluation}

We evaluate quantitatively the motion quality and responsiveness of our system.
Overall, MVAEs can generate high-quality motions and responsive controls that are comparable to existing kinematic motion synthesis methods.

\bfpara{Foot Skating Artifacts}
We use the same measurement as \cite{MANN} to estimate the amount of foot skating during motion, i.e.~$s = d(2 - 2^{h / H})$, where $d$ is the foot displacement and $h$ is the foot height of two consecutive poses.
To account for differences in motion capture data, we use a height threshold of $H = 3.3$cm, which produces an average foot skate of $0.10$ centimeters per frame in the ground truth data, similar to previous work.
\autoref{tab:foot-skate} shows the average foot skate for different MVAE models.
For ease of comparison, the values are presented in centimeters per frame.

\begin{table}
\caption{The average foot skating in the motion capture data, and for MVAEs trained with different KL ($\beta$) and number of experts ($N$).}
\label{tab:foot-skate}
\begin{tabular}{@{}lrrrrr@{}}
\toprule
\textit{cm/frame} & Random & Target & Joystick & Path & Maze\\
\midrule
Motion Capture & 0.10\\
\cmidrule(r{0.5em}){1-2}
$\beta = 0.2, N = 6$ & 0.067 & 0.27 & 0.28 & 0.30 & 0.24\\
$\beta = 0.4, N = 6$ & 0.082 & 0.15 & 0.33 & 0.28 & 0.21\\
$\beta = 0.2, N = 4$ & 0.085 & 0.24 & 0.39 & 0.44 & 0.38\\
\bottomrule
\end{tabular}
\end{table}

\bfpara{Joystick Responsiveness}
We also measure the controller responsiveness under the joystick control task, where a new target direction is randomly sampled every five seconds.
Since the desired heading direction changes instantaneously, we count the number of frames the character takes to reach within five degrees of the target direction.
\autoref{tab:responsiveness} summarizes the result.
The breakdown of the response time by target direction demonstrates the effect of handedness, which we further discuss in~\autoref{sec:handedness}.

\begin{table}
\caption{The average controller responsiveness when performing the joystick control task. The last two columns show the response times for target directions in the left and right half-plane.}
\label{tab:responsiveness}
\begin{tabular}{lrrrr}
\toprule
\textit{seconds} & Overall Time & $\left(0, \pi\right]$ & $\left( \pi, 2\pi \right]$\\
\midrule
$\beta = 0.2, N = 6$ & 1.62 & 1.70 & 1.56\\
$\beta = 0.4, N = 6$ & 1.71 & 1.66 & 1.77\\
$\beta = 0.2, N = 4$ & 1.38 & 1.63 & 1.11\\
\bottomrule
\end{tabular}
\end{table}

\subsection{Ablation on Decoder Architecture}

We use the MoE decoder model because it produces higher motion quality with less visible artifacts.
While a non-mixture model can also produce quality motion, we find it to be less consistent.
In extreme cases, the predicted pose can converge to the mean pose of possible next frames, causing the character to be stuck in the same pose while gliding.
Furthermore, we experimented with the effect of encoder latent dimension size and number of decoder experts for the MoE architecture.
In both cases, we find that the motion quality is not particularly sensitive to the choices, however, the divergent behaviour occurs later with growing latent size and number of experts.
The ablation results are shown in the supplementary video.

\subsection{Using a Noisy Policy to Generate Motion Variations}

\begin{figure}
  \centering
  \includegraphics[width=1\columnwidth]{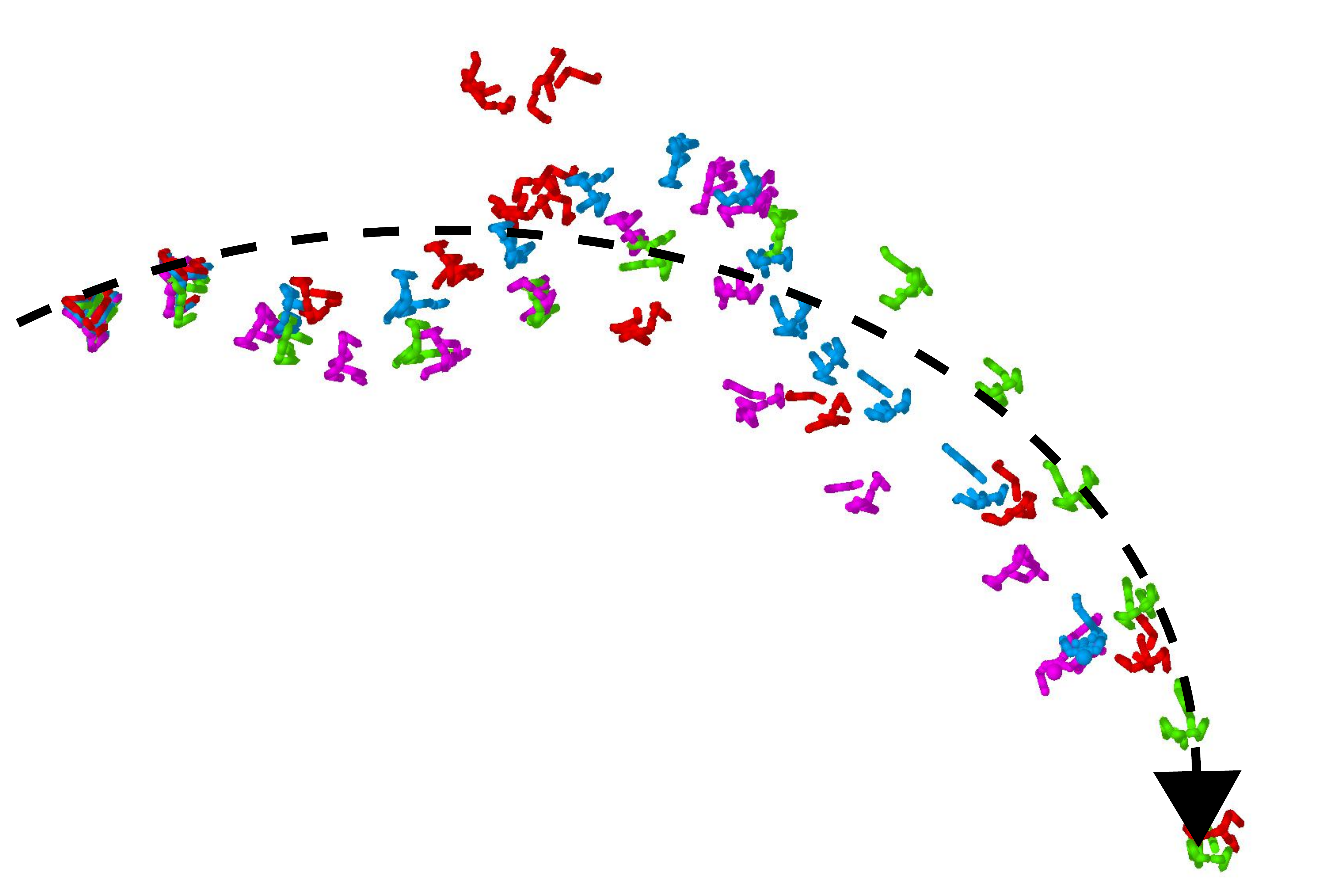}
  \caption{It is possible to generate a variety of motions while still satisfying the overall objective.  
  We simulate four characters with the same initial condition and with the same path following control policy, 
  but we include the selection of random actions with a probability of 0.2 at each time step.
  The controllers are exposed to noise during the exploration stage of RL, and so can recover even if the path deviates at run-time.}
  \label{fig:noisy-controller}
\end{figure}

After RL training, we can create plausible motion variations by sampling around the output of the trained policy.
Specifically, motion variations can be achieved by adding a small amount of noise to the actions at run-time.
As the impact of the noise accumulates over time, 
the trajectories of the simulated characters become visibly distinguishable from one another.
To demonstrate the effect, we simultaneously simulate multiple characters in the Path Follower task.
We set the characters to have the same initial root position and pose.
Since the target progression is the same for all characters, the variations are the result of the noisy decisions at run-time.
\autoref{fig:noisy-controller} shows the variation between individual characters, while there are all still ``on task''.

\subsection{Acyclic Motions}

We can integrate acyclic motions into MVAEs by providing the motion type as an additional condition to the encoder and decoder.
We categorize clips in the motion capture database by their motion types: locomotion, kick, and header.
The category is represented as an one-hot vector and is concatenated with the pose representation to form the new VAE condition.
Although each kick and header clip contains a segment of locomotion leading up to the final action, we find that the MVAE can handle the labelling ambiguity.
The results are shown in the supplementary video.

Since our motion database contains relatively small amounts of kick and header data, it can be difficult for the character to discover these motions during random walk and RL training.
To reduce the sampling complexity, we can explicitly blend the target motion type to trigger a transition.
The blending procedure needs to be fine-tuned to avoid reduced motion quality caused by the condition being out of the training distribution.
This issue is analogous to the future trajectory generation procedure in \cite{PFNN, MANN}.
In practice, we find that forcing the target motion type for 10 frames helps with RL training and does not cause significant motion artifacts.
In the future, we wish to determine better ways to assert
finer control over specific motions, through additional data or data reweighting.

\subsection{Handedness Bias}
\label{sec:handedness}

In our experiments, we observe that the learned task controllers often exhibit a slight preference for right-handedness, 
despite the motion capture database being left-right balanced.
We believe that the handedness bias emerges from exploration and exploitation during RL training.
A policy in training may first discover, by chance, that the right-hand turn motion achieves a 
higher than average reward in some situations.
The probability of sampling this motion is subsequently increased due to the higher return, 
which leads to lower probability of performing a left-hand turn.
In physics-based animations, a typical solution is to enforce symmetry during training of the controllers, 
using one of several possible methods~\cite{SymmetricLoco}.
However, the symmetry enforcement methods all require the action mirror function to be defined, 
which is undefined for our uninterpretable latent action space.

\subsection{Limitations}

Our method has a number of limitations.
The distribution of example data plays a role in determining the likelihoods of the stochastic motion model. 
For example, if right-hand turns greatly outnumber left-hand turns, then this will be reflected in random walks using the model.
Similarly, the final motion connectivity is dependent on the approximate connectivity available in the input data.
For example, our motion dataset contains a moderate amount of walking data (see Fig.~\ref{fig:mocap-distribution}), but
we found it difficult to generate control policies that perform the tasks at a walking pace.
We attribute this to a lack of walking data or to lacking connectivity for that data.

While the MVAE and control policy are conceptually separated in our method,
in practice the learned control policy will not be fully agnostic to the motion data distribution.
This is because of the fundamental nature of the stochastic policies that are at the heart
of on-policy policy-gradient algorithms.  A policy with a mean action that walks straight
will also produce samples that perform other nearby actions in the action space, due 
to the (commonly Gaussian) distribution of the stochastic policy actions. In a world that is
more heavily populated by right turns than left turns, nearby actions are then more likely to turn
right than left.

In general, it can be difficult to attribute a problem to a given portion of our learning pipeline.
Problems may arise because of any of: 
(i) missing data or heavily biased data; 
(ii) MVAE design, including hyperparameters; and 
(iii) control policy design, including reward functions, hyperparameters, and how the character senses its environment.
For example, we found it difficult to generate control policies that can navigate in more tightly-constrained environments.

\section{Conclusions}

In this paper, we have presented a VAE-based approach for motion synthesis.
We show that VAEs are a viable learned stochastic model for motion dynamics,
and can produce robust, high-quality, long-term motion predictions even for
a simple memoryless first-order autoregressive model. 
Reinforcement learning can then be used to learn control policies on top of the
learned motion VAE, using the stochastic latent variable as the action space.
Unlike direct-prediction approaches, which directly learn final task-relevant motion predictions from example data, the learning of the task is separated from
the learning of the dynamics, which allows multiple control policies to
be learned using the same motion model. 

For future work, we wish to explore the best ways for artists to be able to exert control over the MVAE and the learned control policies.
\edit{Tools for identifying gaps and adding connectivity to the underlying example data will improve the design efficiency.}
We also wish to test the model on much larger motion datasets, which should allow for a variety of non-locomotion tasks.
Including more environment context in the MVAE is an important direction, given that many tasks involve controlled interaction with the world.
It may be possible to extend MVAEs to interesting multi-agent settings.

\begin{acks}
We thank Sebastian Starke for the stimulating discussions and his help on high-quality rendering, Elly Akhoundi for sharing her insights on VAEs, Matteo Loddo for providing the character assets, and Paul McComas for his support on this project.
H.L.~thanks Wil Kao for his research tips, which are often as motivating as seeing learning curves finally converging.
This work was supported by Mitacs through the Mitacs Accelerate program.
\end{acks}

\bibliographystyle{ACM-Reference-Format}
\bibliography{bibliography}


\end{document}